\documentclass[conference]{IEEEtran}
\IEEEoverridecommandlockouts
\usepackage{cite}
\usepackage{amsmath,amssymb,amsfonts}
\usepackage{algorithmic}
\usepackage{graphicx}
\usepackage{textcomp}
\usepackage{xcolor}
\usepackage{ulem}
\usepackage{color}
\usepackage{subfigure}
\usepackage{ulem}
\usepackage{xcolor}
\usepackage{float}
\usepackage{url}
\usepackage[T1]{fontenc}
\usepackage[utf8]{inputenc}
\usepackage{authblk}
\usepackage{lipsum}
\def\BibTeX{{\rm B\kern-.05em{\sc i\kern-.025em b}\kern-.08em
    T\kern-.1667em\lower.7ex\hbox{E}\kern-.125emX}}
\begin{document}

\title{HeteroMorpheus: Universal Control Based on Morphological Heterogeneity Modeling\\}

\author[*]{YiFan Hao}
\author[*]{Yang Yang}
\author[**]{Junru Song}
\author[*]{Wei Peng}
\author[*]{Weien Zhou}
\author[*]{Tingsong Jiang}
\author[*]{Wen Yao}
\affil[*]{Defense Innovation Institute, Chinese Academy of Millitary Science, Beijing, China}
\affil[**]{Institute of Statistics and Big Data, Renmin University of China, Beijing, China}
\renewcommand\Authands{ and }
\maketitle

\begin{abstract}
In the field of robotic control, designing individual controllers for each robot leads to high computational costs. Universal control policies, applicable across diverse robot morphologies, promise to mitigate this challenge. Predominantly, models based on Graph Neural Networks (GNN) and Transformers are employed, owing to their effectiveness in capturing relational dynamics across a robot's limbs. However, these models typically employ homogeneous graph structures that overlook the functional diversity of different limbs. To bridge this gap, we introduce HeteroMorpheus, a novel method based on heterogeneous graph Transformer. This method uniquely addresses limb heterogeneity, fostering better representation of robot dynamics of various morphologies. Through extensive experiments we demonstrate the superiority of HeteroMorpheus against state-of-the-art methods in the capability of policy generalization, including zero-shot generalization and sample-efficient transfer to unfamiliar robot morphologies.
\end{abstract}
\renewcommand{\thefootnote}{\fnsymbol{footnote}}
\footnote[1]{Also with Intelligent Game and Decision Laboratory, Chinese Academy of Millitary Science.} 
\let\thefootnote\relax\footnotetext{Corresponding author: Wen Yao (wendy0782@126.com)}

\section{Introduction}
The integration of reinforcement learning with robotic control tasks has led to significant advancements\cite{singh2022reinforcement}, reaching expert-level performance in various domains without the need for the expert expertise required by traditional robotic control methods\cite{r27}. However, this progress is tempered by the limitation that the policy of each robot is required to independently learn from scratch, a process that escalates computational costs as the number of robots increases. To address this challenge, the universal control policy emerges as a promising solution. By conceptualizing the training of robots with varying morphologies as a multi-task reinforcement learning problem, the universal controllers propose to develop a singular policy model applicable across different morphologies, not only enhances learning efficiency of initial forms but also fosters generalization capability to unseen ones.
\par
The role of morphological information in the effective implementation of universal controllers remains a subject of considerable debate. Wang et al. \cite{r5} and Huang et al. \cite{r6} pioneered the use of tree structure to universally represent robots of various morphologies, employing Graph Neural Networks (GNNs) to facilitate information aggregation along heuristically designed paths. They contend that the morphological information harnessed for message passing is crucial for global coordination, thereby playing a pivotal role in the successful adaptation to different morphologies. Contrarily, Kurin et al. \cite{r8} challenge this viewpoint, highlighting the 'over-smoothing' problem in GNNs which can lead to significant information loss from remote nodes during multi-hop communications. They argue that the detriments arising from this loss might overshadow the benefits derived from the graph structure in GNNs. As a solution, they advocate for a morphology-agnostic approach using Transformers, where the self-attention mechanism permits unrestricted information flow and attention computation between nodes, akin to applying GNNs on a fully-connected graph. This method forgoes the reliance on the morphological structure for message passing, consequently yielding notable enhancements in control performance and generalization. However, this perspective is not unanimous. Hong et al. \cite{r11} assert that incorporating morphological information into state observations, including positional and relational embeddings, introduces a beneficial inductive bias that can significantly outperform the morphology-agnostic Transformer model. Parallel findings by Gupta et al. \cite{r17} and Brandon et al. \cite{r24} further corroborate this insight.
\par
However, current methods uniformly model robot morphology in graph structures, treating all limb connections as homogeneous nodes and edges, with Transformer approaches further neglecting edge influences. This homogenization fails to account for the diverse types and functionalities of limbs, potentially hindering the exploitation of their distinct functions. Moreover, treating edges uniformly and adopting a fully connected design negatively impacts model control by conflating all node information through the same mapping process, without adequately distinguishing between nodes based on type and location. Such an approach also compromises the robot's structural integrity by assuming equal relevance for all limb connections, contrary to findings by Dong et al. \cite{r23}, who demonstrate distinct relationships between different limbs. Their research reveals organ-specific limb synergies, indicating significant variance in information transmission within and across such organ systems. This analogy extends to human limb motion, where muscle coordination varies markedly between different joint combinations. Thus, past methodological oversimplifications may impede effective limb control and restrict the model's capacity to devise optimal universal control strategies.
\par
In this paper, we present HeteroMorpheus, a novel approach utilizing the Heterogeneous Graph Transformer (HGT) \cite{r12} to model robots as heterogeneous graphs. This approach enables a nuanced representation of limb diversity and interactions. Evaluated in Evogym \cite{r13}, a benchmark environment encompassing a range of motion and manipulation tasks for soft robotic control, HeteroMorpheus demonstrates superior task performance and morphological adaptability over existing techniques. Moreover, it also exhibits remarkable generalization capabilities and efficiency in fine-tuning for previously unseen morphologies.

\section{RELATED WORK}
The concept of universal control policy was formally introduced by Wang et al. \cite{r5}, where they formulate the problem as multi-task reinforcement learning and represent robot morphologies as kinematic graphs. They employ GNNs to foster generalization across diverse graphs, each with distinct nodes and edges, effectively resolving the issue of varied observation and action spaces across different robots. Following this, Huang et al. \cite{r6} propose Shared Modular Policies (SMP) that view robots as assemblies of functional modules. The SMP adopts GNNs to faciliate the exchange of proprioceptive observations among modules through strategically designed top-down and bottom-up information pathways, enabling the control of different robots within a same policy network. Despite the advantages of GNN-based approaches, they suffer from performance degradation as the number of robot limbs increases. This is due to the loss of information caused by multi-hop communication during message passing, making it difficult to fully transmit information from distant nodes.

\par
Instead of utilizing kinematic graph to define message-passing schemes, Kurin et al. \cite{r8} propose the hypothesis that any benefit GNNs can extract from the morphological domain knowledge encoded in the graph is outweighted by the difficulty that the graph creates for message passing. Then they propose Amorpheus, a universal controller that completely ignores morphological information and utlize Transformer \cite{r10} to build the policy network, taking message passing process as attentional aggregation over fully connected graphs. However, Hong et al. \cite{r11} in a following study argue that morphological information is still very usefull. They extract graphical features from robot kinematic tree and encode them into attentional aggregation, substantially improving the performance of Amorpheus.

\par
Unlike adding robot's graphical features to the attention aggregation process, Gupta et al. \cite{r17} believe that robot morphology is just another modality on which they can condition the output of a Transformer. Therefore, they choose to expand the input of the control model, incorporating morphology information as part of the observed input to the model. They also add learnable position encodings, which contribute to excellent performance. On the other hand, Zheng et al. \cite{r18}  argue that whether adding a morphology-aware positional encoding to the state representation \cite{r11} or feeding morphology context as an additional input to the policy network \cite{r17}, they are effectively equivalent to just adding a context-conditioned bias term to a specific layer in the network, which may not be sufficient to give the control model enough capacity to represent policies that can control different morphologies. Therefore, Zheng et al. \cite{r18} propose to use hypernetworks \cite{r19} to generate morphology-dependent control parameters based on MetaMorph, aiming to better model the dependency of the control policy on the robot morphology.
\par
Some people have raised objections to the default robot configuration definition when designing generic control strategies. Dong et al. \cite{r23} draw inspiration from the human central nervous system's ability to effortlessly control hundreds of muscles through muscle synergies. They incorporated muscle synergies into their universal control policy. Based on the assumption that muscles in coordination have similar functions and similar morphological context, they consider limbs that are close to each other and have similar functions as having coordinated action and partition them into the same organ, further learning the structure of organs for the robot's morphology. Their method learns and outputs the action selection for each synergy (i.e., organ). The partitioning of limbs into different organs for coordinated action design highlights the heterogeneous relationship of mutual influence among robot limbs, which provides us with inspiration. On the other hand, Trabucco et al. \cite{r24} argue that the manual design of robots as graph structures and the assumption that each limb has corresponding sensors and actuators are not accurate for defining morphology. Therefore, they no longer assume a one-to-one correspondence between observation information, action signals, and the composition module of the robot. Instead, they adopt a data-driven approach to automatically learn the representation of the robot's morphology through the reinforcement learning process for training universal control policy.

\par
Unlike existing methods that transmit morphology information to the controller through feature concatenation or position encoding, our approach focuses on improving the design of the message passing scheme defined by morphology information. By heterogeneously modeling nodes and edge relationships, the control model can have a more comprehensive understanding and mastery of the morphology of the robot and the coordination among its limbs. This enables effective enhancement of the control capabilities for the robot, as well as the generalization ability to unknown morphologies.

\section{BACKGROUND}
This section introduces the benchmark testing environment and the problem of learning a universal control policy for a group of robots with different morphologies.

\subsection{Evolution Gym}

We choose Evolution Gym (Evogym) \cite{r13}, a specialized simulation platform for voxel-based soft robots (VSRs), as our evaluation environment. Unlike rigid-body simulations, VSRs are normally arranged in grid structure, offering expansive design possibilities compared to rigid robots. Existing tree-structured approaches like NerveNet \cite{r5} and SMP \cite{r6} struggle with VSR morphologies due to inadequate structural representations and the risk of information loss over extended message passing paths. Moreover, methods enabling unrestricted information flow, such as Amorpheus \cite{r8} and MetaMorph \cite{r17}, lead to increased computational cost. We introduce HeteroMorpheus to overcome these challenges. The HeteroMorpheus precisely captures heterogeneous relations among voxels and simply aggregates information from immediate neighboring voxles, significantly outperforming state-of-the-art (SOTA) methods. 

\begin{figure*}[t]
\centering
\includegraphics[width=1\linewidth]{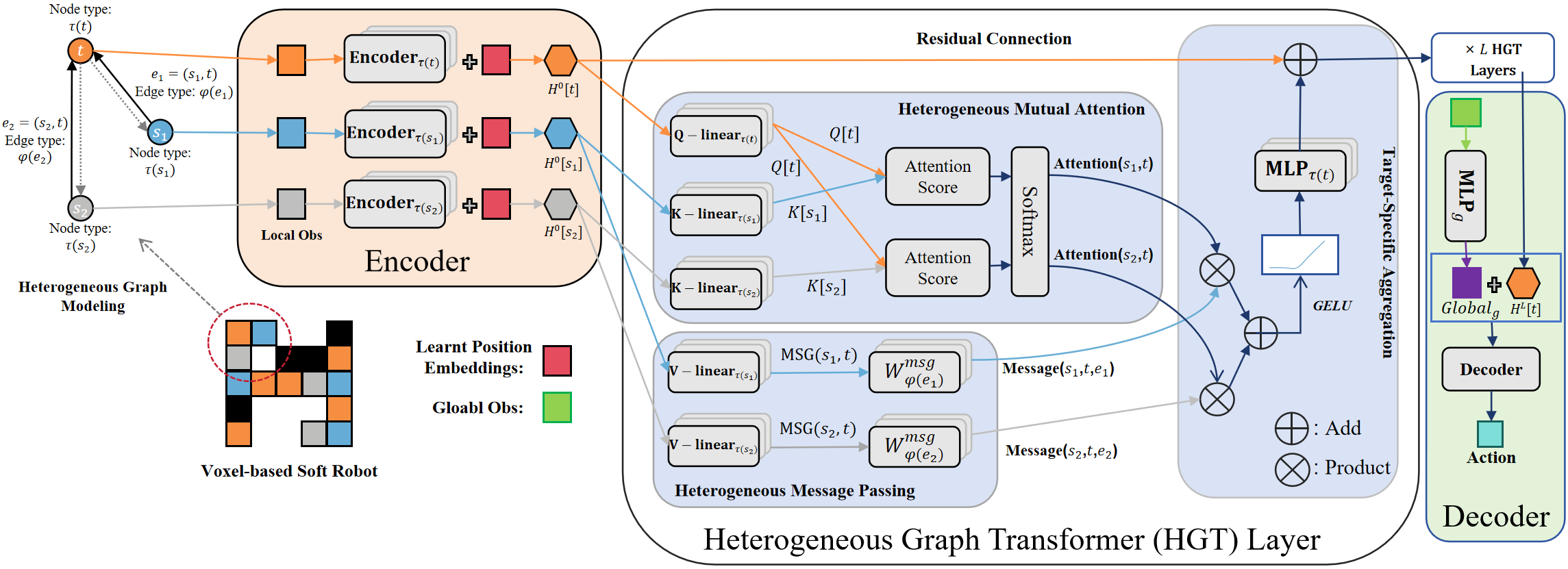}
\caption{{\bf~The Overall Architecture of HeteroMorpheus.} We illustrate the action generation process for a voxel node in the robot. The local observations of the target node $t$ and its neighboring nodes $s_1$ and $s_2$ are linearly embedded based on their corresponding node types. The embedded vectors are then added to the learnt positional embedding as the initial features of the node, which are subsequently fed into the HGT for heterogeneous attention calculation, message passing, and aggregation. The output of HGT is concatenated with the linear embedding of the global observation and passed to the Decoder to generate a distribution over actions.}
\label{fig:picture001}
\end{figure*}

\subsection{Problem Formulation}
Learning a universal controller on a group of $K$ robots with different morphologies generated from a modular robot design space is a multitask reinforcement learning problem. For each robot $k$ can be represented as a heterogeneous directed graph with distinguishable node and edge types: $G=(V,E,U,P)$, Each node $v^{i}\in V$ represents a voxel block of the robot, where the voxel type corresponds to the node type. Directed edge $e^{i j}\in E$ represents the neighbourhood relationship from node $v^{i}$ to $v^{j}$.  Each node and each edge are associated with their type mapping functions, respectively, represented as $\tau(v)\colon V\to U$ and $\varphi(e)\colon E\to P$. 
\par
For each robot $k$, the control process of the robot can be viewed as a Markov decision process consisting of tuples $(M_{k},S_{k},A_{k},T_{k},R_{k})$, where $S_{k},A_{k},T_{k},R_{k}$ represent the state space, action space, state transition function, and reward function, respectively. The set $M_{k}$ contains the morphological information of the robot $k$. At each time step $t$, the robot controller receives the observation information $s_{k,t}$ and the morphological information $M_{k}$. It outputs the action $a_{k,t}$ according to the policy $\pi$ and receives the reward $r_{k,t}$. The training objective is for the controller to learn a universal policy $\pi_{\theta}(a_{k,t}|s_{k,t},M_{k})$ that maximizes the average reward for all morphologies, i.e $m a x_{\theta}[\frac{1}{K}\Sigma_{k=1}^{K}{\Sigma}_{t=0}^{H}\,r_{k,t}]$, $H$ is the task horizon for all different robots. We use the PPO algorithm \cite{r22} to optimize this objective.

\section{HeteroMorpheus}

In order to further leverage the morphology of robots, we have re-examined the message passing scheme defined by morphology information and combined it with heterogeneous graph Transformer. By modeling the types of nodes and connections between nodes differently and restricting each node to only compute attention and directly message passing with its neighboring nodes, we have learned a universal control policy that can better understand and apply robot morphology information. Our method improves the policy network of MetaMorph by using HGT to provide the ability of heterogeneous modeling for the model. We have implemented HGT by rewriting it using matrix operations to ensure that the model can process data in parallel, thereby improving training efficiency. The proposed method framework is shown in Fig \ref{fig:picture001}, can be divided into three parts: an encoder consisting of linear layers, a heterogeneous graph Transformer for message passing, and a decoder also consisting of linear layers. The encoder and decoder independently process information from each node. Additionally, we have added learnable position embeddings in the control model to enhance its performance.

\subsection{Encoder}
The policy model receives the morphology information $M_{k}$ of the $k$-th robot (including a 2D voxel matrix that records the types of voxels and a connection matrix that describes the geometric structure) and the observation information $S_{k}$ (which can be divided into local observation information $s_k^p$ and global observation information $s_k^g$, such as the centroid speed or the block position in the manipulation task). $s_k^p$ is a one-dimensional sequence formed by concatenating the coordinate information of the robot's voxel blocks in the order of traversing from left to right and top to bottom, it is used as the initial feature information of nodes, together with the additional positional embedding, for subsequent attention calculation and message passing.

\par
After $s_k^p$ is input to the policy model, it is first necessary to divide the local observation information according to the types of nodes contained in the morphology information. It should be noted that the morphology information does not participate in the subsequent node feature updating process. It only serves to provide the node types and the connection relationships between nodes, thus constructing the heterogeneous graph structure. The actual computation involves the observation information $S_{k}$ transmitted from the environment. The divided node feature information is independently input to the Encoder.
\par
In the universal control policies such as Amorpheus and MetaMorph, which use Transformers, the distinction between node types is not considered. Therefore, in the encoder, the embedding of the observed information for any node is mapped using the same linear mapping. However, we choose to learn an independent linear mapping for each type of node, so as to differentiate the distribution of node type features. The computation formula is as follows:
\begin{equation}
    H^{0}(x)=\mathrm{\bf~Encoder}_{\tau(x)}\bigl(s_{x\in N_k}^{p}\bigr)+W_{p o s} 
\end{equation}
$\mathrm{\bf~Encoder}_{\tau\left(x\right)}$  is a linear network related to the type of node, $N_k$ represents all the nodes in robot $k$, $s_{x\in N_k}^{p}\in\mathbb{R}^{d}$ is the feature information corresponding to node $x$, $\tau\left(x\right)$ represents the type of node $x$, $d$ is the feature dimension of the node. $\mathrm{\bf~Encoder}_{\tau\left(x\right)}$  projects the $\tau\left(x\right)$-type node into the corresponding embedding vector. In addition, we add a learnable one-dimensional position embedding $W_{p o s}$ to the embedded node feature, which is used to automatically learn the position information. Finally, we obtain the output $H^{0}(x)\in\mathbb{R}^{d i m}$ of the Encoder, where $dim$ is the feature dimension of the node after linear mapping.

\subsection{Heterogeneous Graph Transformer}
For the key processing steps, similar to Transformer, HGT also stacks $L$ layers. We represent the output of node x in the $l$-th layer of HGT as $H^l(x)$. The processing of node feature information in HGT can be divided into three parts: heterogeneous mutual attention, heterogeneous message passing, and target-specific information aggregation. The actual processing can be summarized as follows: given a target node $t$ and all its source nodes (neighbor nodes) $s\in N(t)$, different linear mappings are selected based on the types of nodes to compute attention, and message passing takes into account the types of neighbors and the edge relationships between nodes, in order to complete the feature update of the target node. The following will be explained step by step.

\textbf{Heterogeneous Mutual Attention.} Firstly, we calculate the heterogeneous multi-head attention. We also adopt the multi-head attention mechanism, with the number of attention heads being $h$. The formula for calculating the multi-head attention between each target node and its source nodes is as follows:
\begin{equation}
    Q^{i}(t)=\ \mathrm{\bf~Q-linear}_{\tau(t)}^{i}\left(H^{l-1}(t)\right),\qquad l=1\ldots L
\end{equation}
\begin{equation}
    K^{i}(s)=\ \mathrm{\bf~K-linear}_{\tau(s)}^{i}\left(H^{l-1}(s)\right),\qquad l=1\ldots L   
\end{equation}
\begin{equation}
    \mathrm{Att}_{\mathrm{head}}^{i}(s,t)=(K^{i}(s)\cdot Q^{i}(t)^{T})/\sqrt{d
i
m/h}, i=1\ldots h
\end{equation}
\begin{equation}
    \mathrm{\bf~Attention}\left(s,t\right)=softmax_{ \forall s\in N(t)}||_{i\in\lbrack1,h\rbrack}\mathrm{Att}^i_{\mathrm{head}\left(s,t\right)}
\end{equation}
\par
Similarly to the approach in the previous Encoder, we maintain a set of linear projections for each node type to calculate the key, query, and value separately. For the $i$-th attention head, $\mathrm{Att}^i_{\mathrm{head}\left(s,t\right)}$, we project the feature information of the target node $t$ with node type $\tau(t)$ into a query vector $Q^{i}(t)$ using the linear projection $\mathrm{\bf~Q-linear}_{\tau(t)}^{i}$. We also project the feature information of the source node $s$ into a key vector $K^{i}(s)$ using $\mathrm{\bf~K-linear}_{\tau(s)}^{i}$. Then, we take the dot product of the query and key to obtain the attention score of the target node $t$ towards the source node $s$.
\par
It is important to note that unlike the approach used in previous applications of Transformer, where each node calculates attention with all other nodes, our method only considers the direct neighbors in the four surrounding directions (up, down, left, right) for each node. This allows us to avoid directly receiving information from nodes located further away in the robot structure. Instead, we update the information of neighboring nodes through the stacked network layers, implicitly allowing the target node to gradually aggregate information from distant nodes.
\par
Finally, concatenate the results of $h$ attention heads to obtain the attention vector $\mathrm{\bf~Attention}\left(s,t\right)$ for each target-source node pair. For each target node $t$, aggregate all the relevant attention scores from its neighbors $N(t)$ and apply softmax processing to ensure that the sum of attention scores for all $\sum_{\forall s\in \mathrm{N}(t)}\mathrm{\bf~Attention}(s,t)$ is equal to 1.

\textbf{Heterogeneous Message Passing.} The source node needs to transmit its own information to the target node for the final information aggregation. Similar to the computation of keys and queries, different mappings are applied to each type of node, and the information from different heads $\mathrm{MSG}_{\mathrm{head}}^{i}$ are concatenated at the end. It should be noted that, to further ensure that the model can effectively utilize morphological information, unlike the design in methods such as Amorpheus and MetaMorph that apply Transformer and assume arbitrary nodes are fully connected, our approach will additionally consider the types of edges involved in message passing process. The concatenated information is multiplied by the edge relationship weight matrix $W_{\varphi(e)}^{m s g}$ corresponding to the edge type between node $s$ and node $t$, obtain the final information transmitted from node $s$ to node $t$. 
\begin{equation}
    \mathrm{MSG}_{\mathrm{head}}^{i}(s,t)=\mathrm{\bf~V-linear}_{\mathrm{\tau(s)}}^{i}\left(H^{1-1}(s)\right)
\end{equation}
\begin{equation}
\mathrm{\bf~Message}(s,t,e)=W_{\varphi(e)}^{msg}\cdot||_{i\in\lbrack1,h\rbrack}\mathrm{MSG}^i_{\mathrm{head}\left(s,t\right)}
\end{equation}
Where $\varphi(e)$ represents the edge relationship type between two adjacent nodes. $\mathrm{\bf~Message}(s,t,e)$ is the actual information transmitted by neighboring nodes.

\textbf{Target-Specific Aggregation.} After the calculations of heterogeneous attention and heterogeneous message passing in the previous two steps, it is necessary to aggregate the information from source nodes with different feature distributions to the target node. The importance of information from each node is measured by attention scores. This is done by multiplying the attention matrix as the weight matrix with the message matrix to obtain an updated vector ${\overline{{H^{l-1}}}}(t)$. Finally, an activation function is applied to this vector and mapped back to the specific type of distribution belonging to the target node, a set of linear projection $\mathrm{MSG}_{\tau(t)}$ related to node types is used for this purpose. Additionally, to avoid the loss of self-feature information of nodes, residual connections are employed. The specific formula is as follows:
\begin{equation}
    {\overline{{H^{l-1}}}}(t)=\oplus_{\forall s\in N(t)}(\mathrm{\bf~Attention}(s,t)\cdot \mathrm{\bf~Message}(s,t,e))
\end{equation}
\begin{equation}
    H^{l}(t)=\mathrm{\bf~MLP}_{\tau(t)}(\sigma(\overline{{{H^{l-1}}}}(t)))+H^{l-1}(t),\qquad
l=1\ldots L
\end{equation}
\par
Exactly, we obtain the output $ H^{l}(t)$ of the target node $t$ in the $l$-th HGT layer, which is then used as input for the next layer. The output of the last layer is the final output feature of the target node $t$. The final output features of all nodes are then passed to the Decoder for the final computation.
\subsection{Decoder}
The decoder is a Multiple Layer Perceptron (MLP). It takes two inputs: the node feature vectors obtained from HGT calculations and the state representation $G l o b a l_{k}$ of global sensory information learned through a two-layer linear network $\mathrm{\bf~MLP_g}$. These two parts are concatenated and passed to the Decoder. The output $\mu$ of the Decoder, along with a fixed parameter diagonal covariance matrix $\Sigma$, is used to model the action distribution as a Gaussian distribution, thereby obtaining the policy. The formula is as follows:
\begin{equation}
    G l o b a l_{k}=\mathrm{\bf~M L P_{g}}(s_{k}^{g})
\end{equation}
\begin{equation}
    \mu(S_{k})=\mathrm{\bf~Decoder}(H^{L}(t),G l o b a l_{k})
\end{equation}
\begin{equation}
    \pi_{\theta}(a_{k}|s_{k})=N(\mu(S_{k}),\Sigma)
\end{equation}
Once we obtain the action distribution, we can sample from it to obtain the action signal corresponding to each voxel block.

\section{EXPERIMENTs}
\subsection{Experimental Setup}
\begin{figure}[!htbp]
	\centering  
	\subfigbottomskip=2pt 
	\subfigcapskip=-5pt 
	\subfigure[Walker-v0]{
		\includegraphics[width=0.48\linewidth]{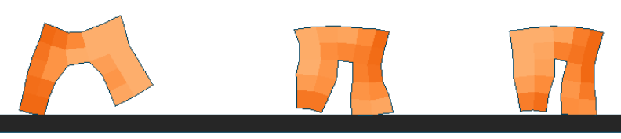}}
	\subfigure[Upstepper-v0]{
		\includegraphics[width=0.48\linewidth]{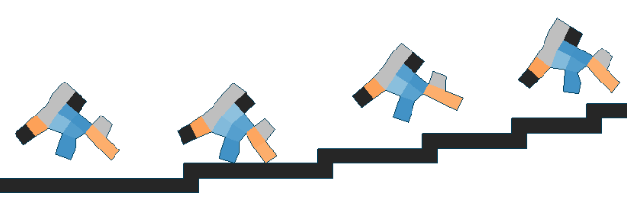}}
	  \\
	\subfigure[Pusher-v0]{
		\includegraphics[width=0.48\linewidth]{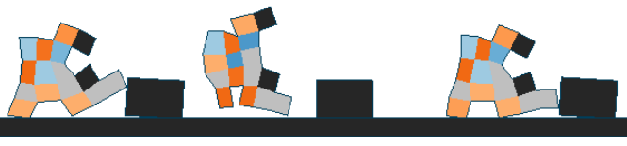}}
	\subfigure[Carrier-v0]{
		\includegraphics[width=0.48\linewidth]{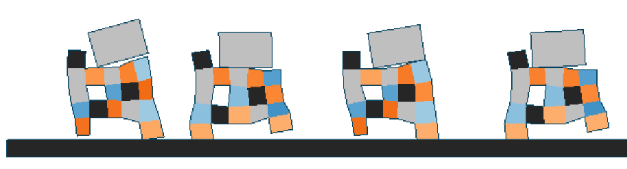}}
  	\subfigure[Catcher-v0]{
		\includegraphics[width=0.48\linewidth]{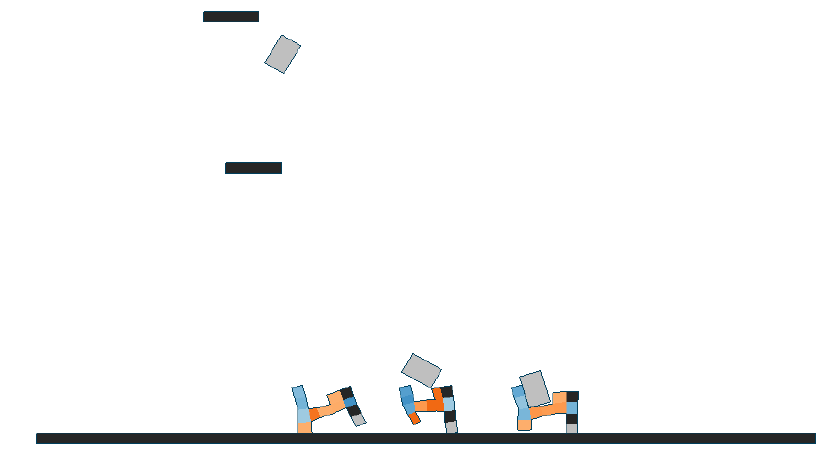}}
	\caption{{\bf~Evogym tasks.} In Walker-v0, the robot's objective is to travel as far as possible on flat terrain. Carrier-v0 involves transporting an object to maximize distance covered. In Pusher-v0, the aim is to push an object to achieve the greatest displacement. UpStepper-v0 presents the challenge of ascending stairs of varying lengths to cover maximum distance. In Catcher-v0, the robot needs to receive an object thrown from a height and in a rotating state, and then carry the object forward as much as possible. All tasks measure progress along the positive x-axis.}
        \label{fig:environment}
\end{figure}

Our experiments encompass five tasks in Evogym, illustrated in Fig \ref{fig:environment}, comprising two locomotion tasks (Walker-v0, UpStepper-v0) and three manipulation tasks (Pusher-v0, Carrier-v0, Catcher-v0).

\par

For each task, We employed the standard PPO algorithm from Evogym to pretrain a set of high-performing robot morphologies, ensuring the exclusion of designs incapable of task completion. We then curated a subset from each high-performing set, maintaining less than $70\%$ voxel count similarity between any two morphologies. This process ensures high diversity of robot morphologies that contributes to proper evaluation of universal control policies. As a result, we collect 25 diversified morphologies for each task, with 15 for training from scratch and 10 for zero-shot generalization and fine-tuning experiments. Within each experiment, all robots are trained in parallel, the PPO algorithm is used for controller optimization, with actor and critic networks are built using HeteroMorpheus. We report comparison results with competitive baselines as well as between different variants of our method. For each method in each task during the training process, we conduct a total of 1000 rounds of parameter training to achieve sufficient training for the control policies. All experimental results are averaged over five independent runs. Our experiments were conducted on a server equipped with 2 AMD EPYC 9754 CPU@2.25GHz and 1 NVIDIA A800 GPU, running Ubuntu 22.04. Our codes, together with an appendix including hyperparameters are available on 
Github(https://github.com/hyf1530588058/HeteroMorpheus).

\subsection{Baseline}
Our method is benchmarked against following competitive baselines:

\begin{itemize}
    \item \textbf{Amorpheus}: Amorpheus employs a standard Transformer to build the actor and critic networks. Adhering to hyperparameters from its original implementation, we set the embedding layer dimension of observation to 128, the number of transformer layers to 3, the number of attention heads to 2, and the attention hidden size to 256.
    \item \textbf{MetaMorph}: MetaMorph introduces global observation, configuration parameter and positional embedding of robots joints to Amorpheus. We follow its original parameter settings to set the embedding layer dimension of local observation to 128,  number of Transformer layers to 5, number of attention head to 1, and the attention hidden size to 1024. The global observation is encoded by a 2-layer MLP with hidden dimensions [64, 64].
\end{itemize}

\begin{figure*}[t]
\centering
\includegraphics[width=0.8\linewidth]{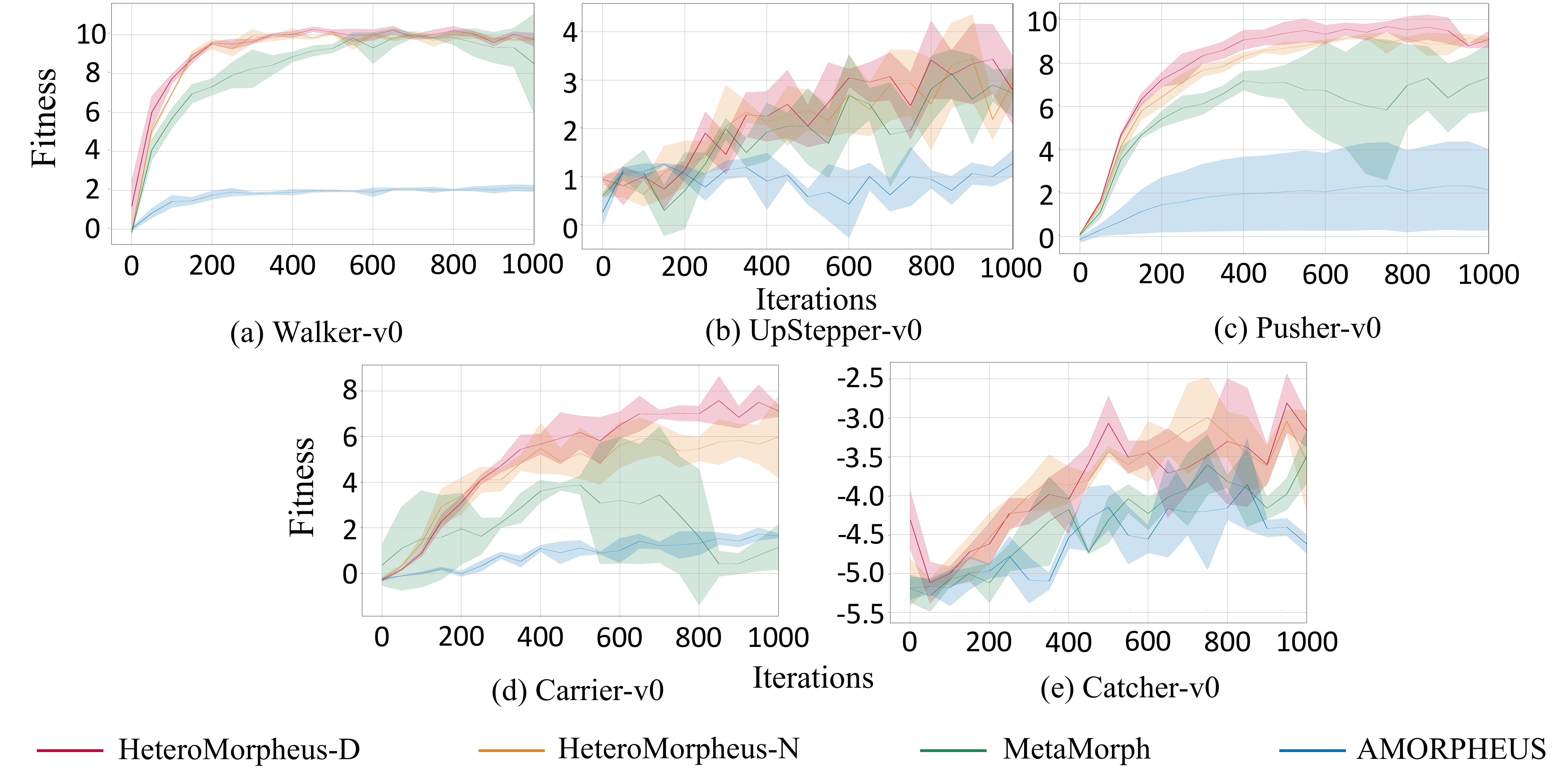}
\caption{{\bf~The training curves of different methods in each environment.} We evaluate on 5 different seeds and plot the mean of average returns over all morphologies. Shaded regions denotes standard deviation.}
\label{fig:train}
\end{figure*}

We designed two variants of HeteroMorpheus: \textbf{HeteroMorpheus-N}, where the edge types are determined by the types of its associated (N)ode pair, excluding empty voxels as target nodes, this leads to $(5*4)$ edge types; \textbf{HeteroMorpheus-D}, where the edge types are determined by the relative (D)irection of neighboring nodes with respect to the target node, with a total of four edge types (up, down, left, right). The embedding dimension of local observations for all variants is 128. The HGT model has 3 layers, 2 attention heads, and the global observation embedding is consistent with the MetaMorph setting. All methods use the same PPO hyperparameters for training.

To ensure fairness, we use the PPO algorithm to optimize each method, adopt parallel training for each task, and unify observation for each method.

\subsection{Training Result}

The experimental results of the training are presented in Figure \ref{fig:train}. It can be observed that both variations of our method outperform the two baseline methods in terms of learning efficiency and the final performance across the four tasks, thus confirming the effectiveness of our approach.

The Amorpheus method, which completely disregards morphological information, exhibits the poorest control performance in different task environments, thereby confirming the significance of morphological information for a universal control policy. The MetaMorph method, which combines position encoding and treats global observation information independently, achieves moderate to above-average performance. However, the homogenization treatment of morphology in MetaMorph results in inadequate and comprehensive utilization of morphological information. Consequently, it slightly lags behind our method in terms of the best performance across various tasks and training stability.

The results demonstrate that incorporating the heterogeneity of morphological structures enables control models to gain a better understanding of morphological information, resulting in improved and more stable control. During manipulation tasks, the requirement to manipulate additional objects with the robot further assesses the control model's ability to effectively utilize the robot's limbs with diverse functions. Thus, our method, which is based on a more precise form modeling approach, has demonstrated significant advantages in tasks such as Carrier-v0, Pusher-v0 and Catcher-v0.

\begin{figure*}[t]
\centering
\includegraphics[width=0.8\linewidth]{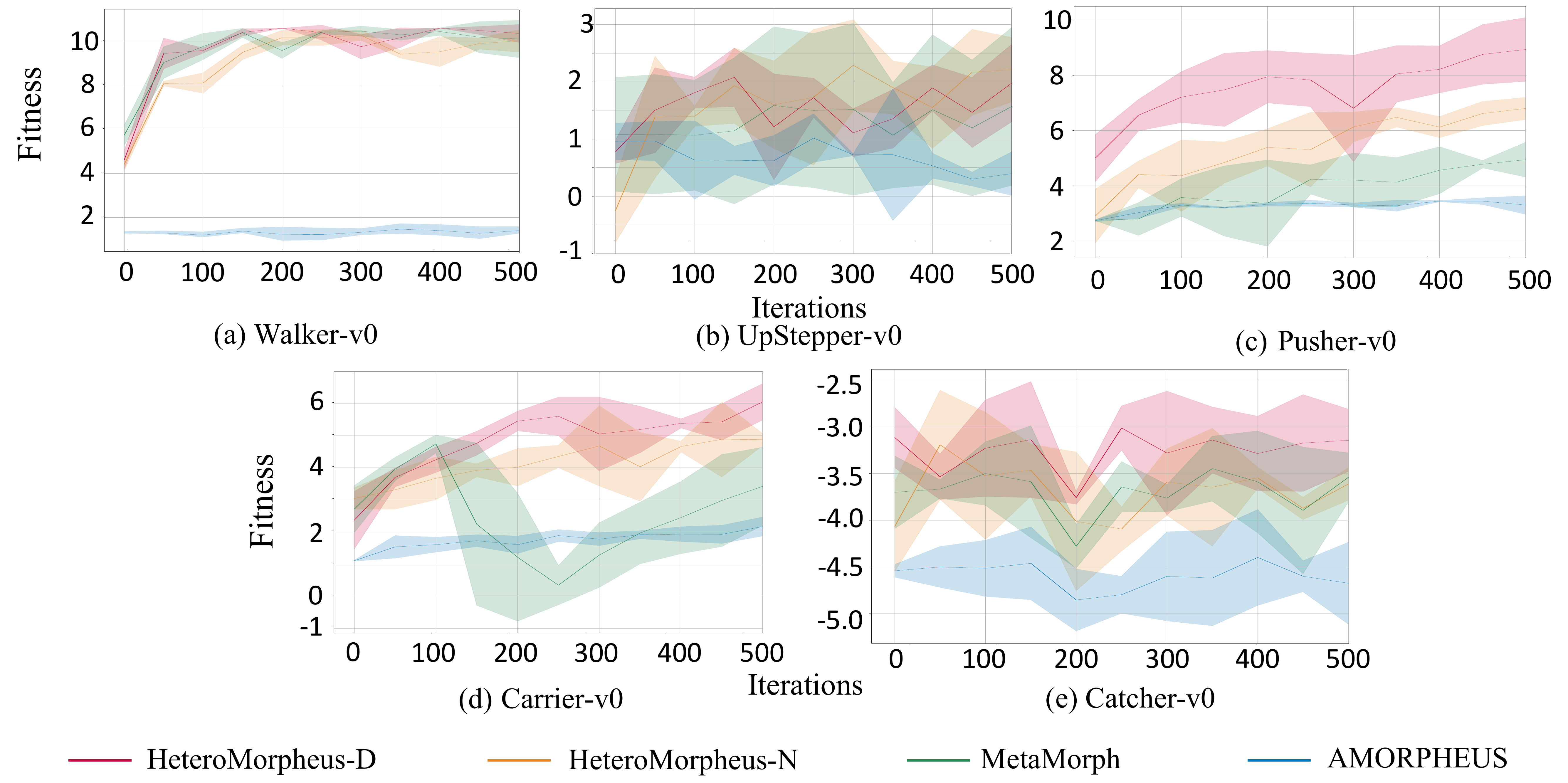}
\caption{{\bf~Comparison of different method's transfer learning performance on an unknown robot morphology set.} The graph presents the results of both zero-shot learning and fine-tuning, obtained through five repeated experiments. Shaded regions denotes standard deviation.}
\label{fig:enter-label}
\end{figure*}

\subsection{Zero-shot and Transfer Learning Result}
In this section, we tested the zero-shot generalization and transfer learning abilities of our approach. In the experimental setup, we directly transferred the pre-trained universal control policy to the unknown sample set for training and testing. The zero-shot and fine tuning results are shown in Fig \ref{fig:enter-label}, and it can be observed that our method achieved the highest generalization scores in the manipulation tasks, which was expected as HeteroMorpheus demonstrated significant advantages over the baseline during training on these three tasks. Models with higher training scores often perform better in zero-shot generalization. 

However, in the simplest Walker-v0 task and the more challenging Upstepper-v0 task,  MetaMorph performed better. When examining the zero-shot scores for each unknown form, it was found that MetaMorph was able to achieve a score of 10 in certain forms in the Walker-v0 task, while the highest zero-shot scores for the two variants of HeteroMorpheus were around 8.5. However, in the subsequent fine-tuning stage, HeteroMorpheus consistently showed higher learning efficiency and a more stable learning process. We speculate that this phenomenon is due to the fact that HeteroMorpheus restricts the information transmission between target nodes and neighboring nodes, relying on the indirect aggregation of neighboring node features to gather information from distant nodes, while MetaMorph allows information transmission between any two nodes, which helps the individual limb nodes quickly obtain information about the entire morphology in unknown samples. When the task is simple enough or too complex, to the extent that different universal controllers achieve similar levels of control ability during the training process, this fully connected design can highlight its advantages. However, as the fine-tuning stage continues and the model gradually gains cognition of unseen morphologies, there is a divergence in the importance of information transmitted by neighboring nodes and non-neighboring nodes. The information from distant nodes may not necessarily help the control of the target node, and the process of relying on the model to autonomously learn to distinguish this importance is too time-consuming. In contrast, HeteroMorpheus can avoid potential interference from non adjacent nodes directly received by nodes through artificial limitations, thus often exhibiting better final performance and more stable learning processes in subsequent fine-tuning stages.
\begin{figure}
    \centering
    \includegraphics[width=1\linewidth]{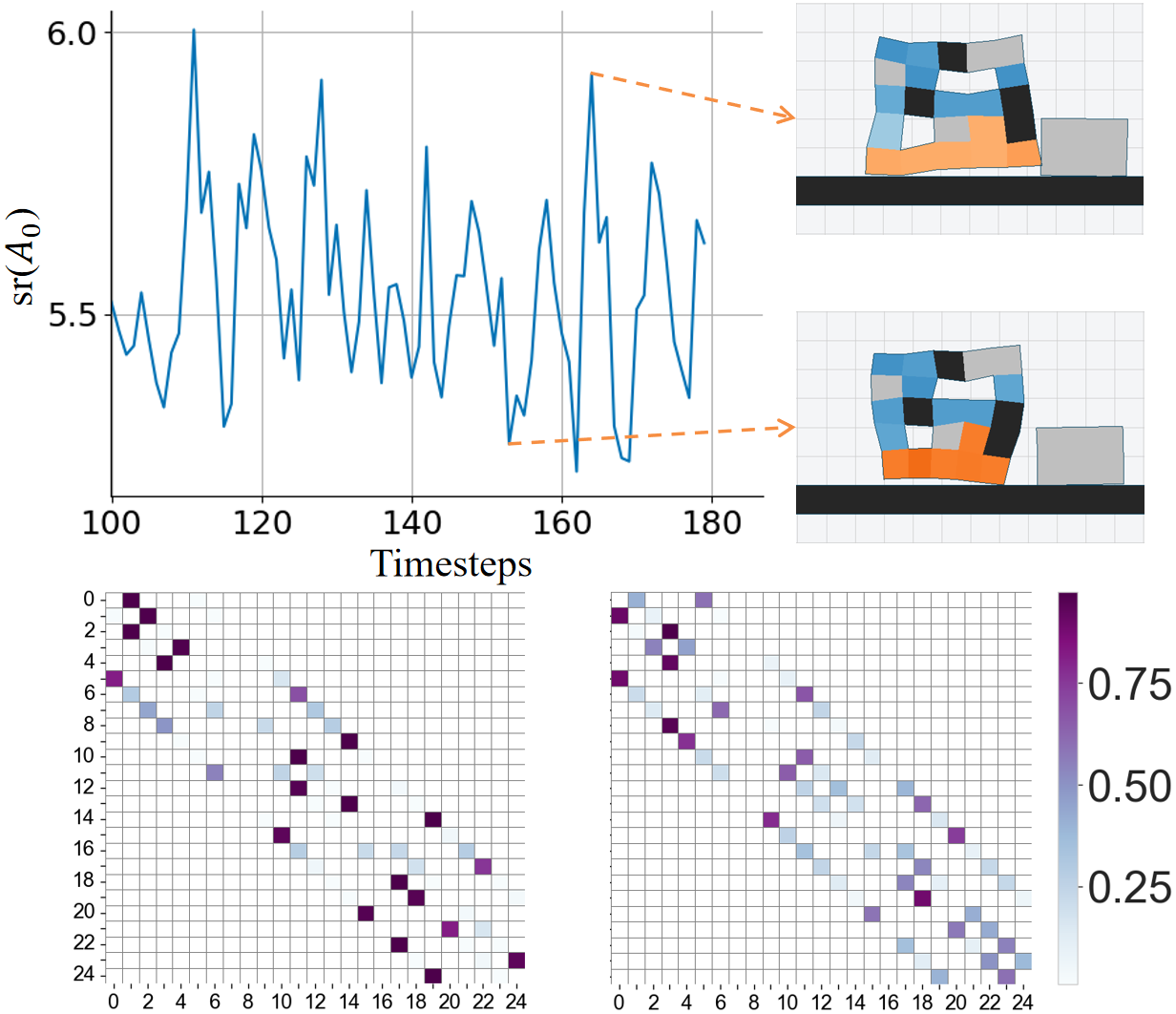}
    \caption{{\bf~Attention Matrix Analysis. }We plot the stable rank of the attention matrix ($sr(A_0)$) for
Pusher-v0, as well as the morphology of the robot and the attention matrix corresponding to the moments when the stabe rank is at its peak and valley. The attention matrix on the left corresponds to the peak, the attention matrix on the right corresponds to the valley. The horizontal and vertical coordinates of the matrix represent the voxel block labels arranged in order from left to right and from top to bottom for the robots.}
    \label{fig:attn}
\end{figure}
\subsection{Attention Matrix Analysis}
To explore the impact of heterogeneous attention mechanisms on universal control policies and their relationship with the robot's motion states, we visualized a segment of the robot's motion process in the Pusher-v0 task and recorded the changes in the robot's attention matrices during this process (As show in Fig \ref{fig:attn}). By plotting the stable rank $sr(A_l)$ of the attention matrices (For attention matrix $A_l$ for layer $l$, the stable rank is defind as $sr(A_l) = \frac{\displaystyle{\sum\sigma_{i}^{2}}}{\displaystyle{\sigma_{m a x}^{2}}}$, $\sigma_{i}$ is the singular values of  $sr(A_l)$), we observed oscillations of the stable rank within a certain range. Examining the robot's motion at the corresponding peaks and valleys of the stable rank, we found a close correlation between the changes in attention matrices and the robot's motion. When the stable rank was at a peak, the robot extended its body to push the block, while at a valley, it contracted its body in preparation for forward motion. The periodic variation of attention matrices controlled the robot's periodic gait.

We also visualized the attention matrices at the peaks and valleys of the graph. It can be observed that during the process of the robot transitioning from contraction to extension, the attention scores of the bottom row, which mainly controls the horizontal actuator voxels responsible for movement, gradually shift from being evenly distributed to all neighboring voxels to focusing only on a specific neighboring voxel. At this moment, the state changes of this neighboring voxel have the greatest impact on the target voxel, demonstrating a closer connection between these two voxels. This close relationship connects multiple voxels together, forming a synergy that can reduce control difficulty. By modeling heterogeneity and considering only messages from neighboring nodes, we can highlight the characteristic differences of different types of nodes and avoid potential noise interference from non-neighboring nodes in the previous fully connected settings. This helps the model to more quickly discover the synergy on the robot, thereby accelerating the learning efficiency of universal control policies.
\section{CONCLUSIONS AND FUTURE WORK}
In this paper, we believe that the homogenized modeling of robot morphology in existing methods diverges from the true form of robots. The homogenization treatment of robot limbs and joints with different functionalities in the control model affects the rational modeling of robot morphology, thereby influencing the control performance of unviersal control methods. To utilize the morphology information of robots more reasonably and accurately, we present a universal control method based on the HGT model. This method models robot morphology as a heterogeneous graph structure, using different neural networks to control limbs and joints of various types. Furthermore, it integrates the morphology-based information transfer mechanism, effectively utilizing the morphology information from a model structure perspective. The effectiveness of our method is validated through experiments on training and transfer learning.

Regarding future work, one direction is to explore the construction of more rational heterogeneous edge relationships. In this paper, we only attempted heterogeneous edge designs based on node types and relative directions. It is also possible to consider combining the two or using new definition methods. Another direction is the selection of neighbor scope. In the analysis of transfer learning experiments, we found that the limited information transmission of adjacent nodes to some extent affected the generalization ability of HeteroMorpheus. In this paper, we consider voxel blocks with shared edges as directly adjacent neighbors. Voxel blocks with shared vertices can also be considered as neighbors. In the study by Dong et al. \cite{r23}, it was found that there may be tighter connections between different limbs of a robot, forming a structure similar to organs. The heterogeneity of organs can also be considered, and limbs belonging to the same organ can be regarded as directly adjacent. Independent heterogeneous message passing schemes can be designed for both within organs and between organs.
\section{Acknowledgements}
This work is funded by the National Key R\&D Program of China under Grant (No. 2019YFA0706401), the National Natural Science Foundation of China(No. 62172014, 62172015, 62272009, 62372008, 62332006), the MOE Project of Key Research Institute of Humanities and Social Sciences (22JJD910002).


\end{document}